\documentclass{article}

\usepackage{microtype}
\usepackage{graphicx}
\usepackage{subcaption}
\usepackage{booktabs} % for professional tables

\usepackage{hyperref}
\usepackage{makecell}

\usepackage[preprint]{icml2026}

\usepackage{amsmath}
\usepackage{amssymb}
\usepackage{mathtools}
\usepackage{amsthm}
\usepackage{tabularx}
\usepackage{array}
\usepackage{multirow} 
\usepackage{threeparttable}

\usepackage[capitalize,noabbrev]{cleveref}

\theoremstyle{plain}

\theoremstyle{definition}

\theoremstyle{remark}

\usepackage{xcolor}
\usepackage{colortbl}
\usepackage{booktabs}
\usepackage{float}
\usepackage{placeins}
\definecolor{lightestpurple}{RGB}{250, 248, 255} 
\definecolor{lightpurple}{RGB}{245, 240, 255}   
\definecolor{mediumpurple}{RGB}{235, 225, 250} 
\definecolor{royalpurple}{RGB}{120, 81, 169}
\usepackage[textsize=tiny]{todonotes}

\DeclareRobustCommand{\eg}{\emph{e.g.}}
\newcommand{\rmnum}[1]{\uppercase\expandafter{\romannumeral #1}}

\newcolumntype{D}[1]{>{\centering\arraybackslash}p{#1}}

\newlength{\imgcropamount}
\newcommand{\croppedimg}[2][\linewidth]{%
    \includegraphics[width=#1, trim=0 \imgcropamount{} 0 \imgcropamount{}, clip]{#2}%
}

\newcommand{\datasetname}{MIRAGE}

\icmltitlerunning{Combined Flicker-banding and Moire Removal for Screen-Captured Images}

\begin{document}

\twocolumn[
  % \icmltitle{Submission and Formatting Instructions for \\
  %   International Conference on Machine Learning (ICML 2026)}
  \icmltitle{Combined Flicker-banding and Moiré Removal \\ for Screen-Captured Images}



  \icmlsetsymbol{equal}{*}

  \begin{icmlauthorlist}
    \icmlauthor{Libo Zhu}{equal,sjtu}
    \icmlauthor{Zihan Zhou}{equal,sjtu}
    \icmlauthor{Zhiyi Zhou}{sjtu}
    \icmlauthor{Yiyang Qu}{sjtu}
    \icmlauthor{Weihang Zhang}{huawei}
    \icmlauthor{Keyu Shi}{huawei}
    \icmlauthor{Yifan Fu}{huawei}
    \icmlauthor{Yulun Zhang}{sjtu}
  \end{icmlauthorlist}

  \icmlaffiliation{sjtu}{Shanghai Jiao Tong University, Shanghai, China}
  \icmlaffiliation{huawei}{Central Media Technology Institute, Huawei, China}

  \icmlcorrespondingauthor{Yulun Zhang}{yulun100@gmail.com}


  \vskip 0.3in
]



\printAffiliationsAndNotice{}  % no special notice (required even if empty)

\begin{abstract}
Capturing display screens with mobile devices has become increasingly common, yet the resulting images often suffer from severe degradations caused by the coexistence of moiré patterns and flicker-banding, leading to significant visual quality degradation. Due to the strong coupling of these two artifacts in real imaging processes, existing methods designed for single degradations fail to generalize to such compound scenarios. In this paper, we present the first systematic study on joint removal of moiré patterns and flicker-banding in screen-captured images, and propose a unified restoration framework, named CLEAR. To support this task, we construct a large-scale dataset containing both moiré patterns and flicker-banding, and introduce an ISP-based flicker simulation pipeline to stabilize model training and expand the degradation distribution. Furthermore, we design a frequency-domain decomposition and re-composition module together with a trajectory alignment loss to enhance the modeling of compound artifacts. Extensive experiments demonstrate that the proposed method consistently outperforms existing image restoration approaches across multiple evaluation metrics, validating its effectiveness in complex real-world scenarios. Our code and dataset will be released at \url{https://github.com/libozhu03/CLEAR}.
\end{abstract}

\vspace{-8mm}
\section{Introduction}
\begin{figure*}[t]
    \centering
    \includegraphics[width=\textwidth]{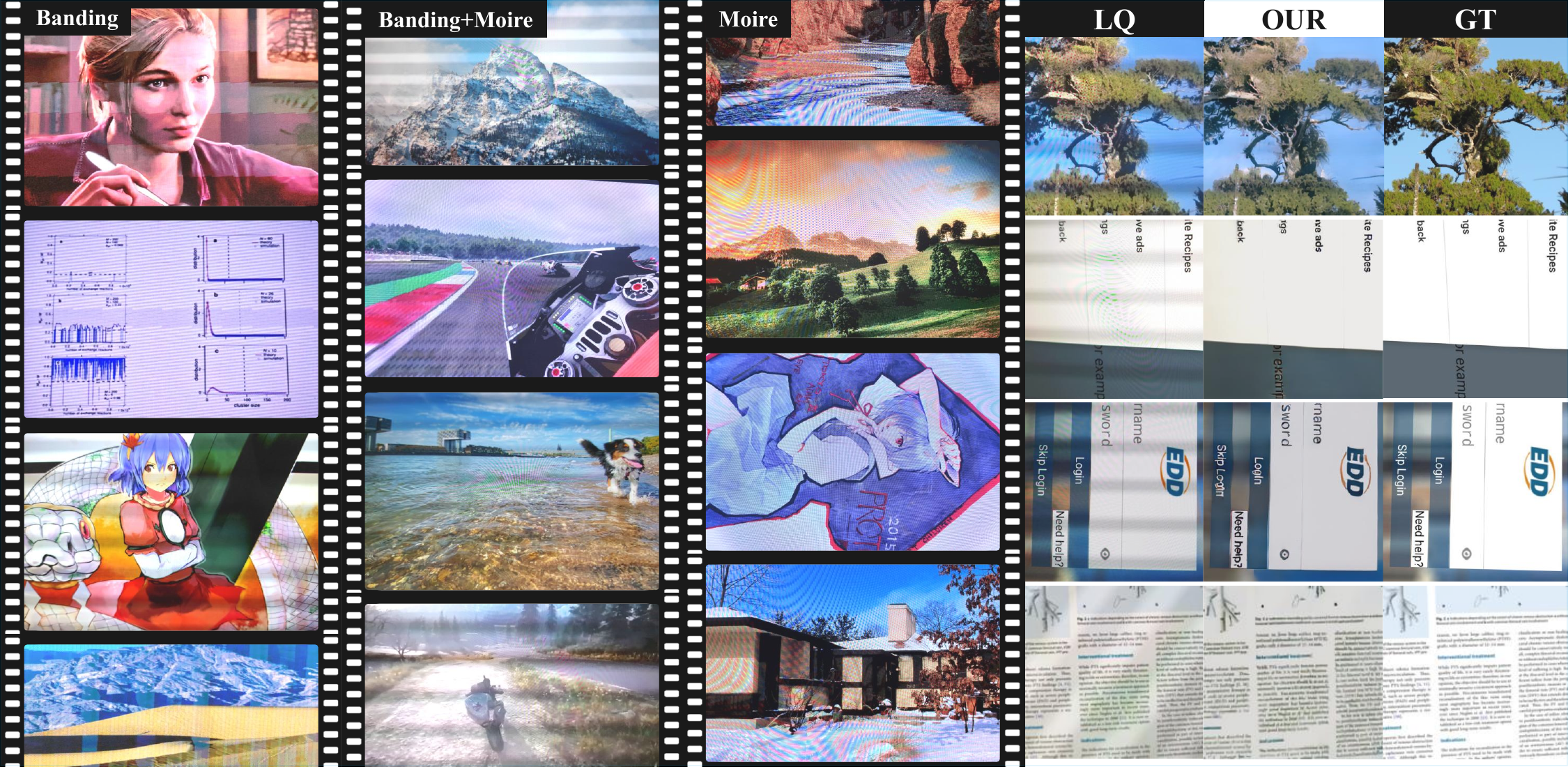}
    \vspace{-5mm}
    \caption{
        Overview of our dataset (\textbf{\datasetname}) and model (\textbf{CLEAR}) results.
        Left: examples of three types of degradations in our dataset: flicker-banding, moiré, and flicker-banding\&moiré.
        Right: the flicker-banding\&moiré removal results of our proposed model on real-world images compared with Low-Quality (LQ) and Ground-Truth (GT) images.
    }
    \vspace{-5mm}
    \label{fig:example of dataset}
\end{figure*}
With the rapid advancement of mobile imaging technology, smartphones have become the dominant tools for daily photography and video recording~\cite{zhou2021image}. Despite continuous improvements in sensors and image processing pipelines, mobile cameras still suffer from non-negligible degradations (\eg, moiré) under certain practical scenarios, which severely degrade the visual quality and usability.

One such scenario arises when capturing display screens, including computer monitors, tablets, and mobile devices. In these cases, the captured images often exhibit structured dark band-like artifacts across the screen region, significantly degrading visual uniformity and obscuring screen content. Given the prevalence of screen-capturing in daily applications, understanding and addressing this degradation remains an important yet underexplored problem.

This phenomenon is commonly known as \textbf{flicker-banding}, which originates from the interaction between rolling shutter image acquisition and pulse width modulation (PWM) based screen brightness control. Rolling shutter sensors expose the image sequentially, resulting in temporal offsets across sensor rows, while PWM controls the screen luminance by rapidly switching pixel intensities over time. When a rolling shutter camera captures a PWM-controlled screen, different sensor rows sample the screen at different temporal phases. Then it leads to spatially structured luminance inconsistencies, performing as banding artifacts.

In screen-captured images, flicker-banding is often accompanied by another well-known degradation: moiré patterns. Moiré artifacts arise from spatial frequency aliasing between the display pixel grid and the camera sensor sampling lattice, typically appearing as repetitive textures or color distortions. While moiré removal has been extensively studied~\cite{2020_NeurlPS,2021_neuro,yu2022towards,mei2025imagedemoireingusingdual,2025_TPAMI}, flicker-banding has been relatively underexplored, largely owing to its temporal characteristics and the complexity of its formation mechanism.  Recent works~\cite{RIFLE} have explored learning-based approaches for only removing banding. 

More importantly, in real-world scenarios, flicker-banding and moiré patterns frequently coexist and interact. However, experiments show that directly cascading flicker and moiré removal models fails when both degradations are present simultaneously. This is because the two artifacts are strongly coupled in both spatial and frequency domains. Flicker-banding alters the global luminance distributions, affecting moiré statistics, while high-frequency moiré structures interfere with banding detection and suppression. These observations highlight the need for a unified framework capable of jointly removing flicker-banding and moiré artifacts.

We encountered the following challenges in the research. \emph{\textbf{\rmnum{1}, Lack of related dataset.}} Existing datasets typically focus on a single type of degradation and lack samples containing both flicker-banding and moiré artifacts, making them unsuitable for joint restoration. \emph{\textbf{\rmnum{2}, Severe structual and perceptual degradation.}} Images affected by both degradations often exhibit extremely poor visual quality, with diverse banding patterns, severe structural degradation, and great color corruption. The image restoration is extremely challenging, and existing models don't demonstrate satisfactory performance. \emph{\textbf{\rmnum{3}, Unstable training on real-world data.}} Due to the diversity of degeneration, directly training deep models on real captured data is highly unstable, frequently resulting in non-convergence or training collapse. 

% It indicates that directly training only on the real-world flicker-banidng\&moiré dataset is not the optimal solution.

To address these challenges, we construct a large-scale dataset: \textbf{M}oiré and fl\textbf{I}cke\textbf{R} in screen im\textbf{A}\textbf{G}\textbf{E} capture (\textbf{\datasetname}), containing both moiré patterns and flicker-banding by capturing screen-displayed images under diverse devices, displays, and shooting conditions. Furthermore, we propose a restoration model named \textbf{C}ombined f\textbf{L}ick\textbf{E}r \textbf{A}nd moiré \textbf{R}emoval for Screen-Captured Images (\textbf{CLEAR}). 

To optimise the distribution of training data and expand the diversity of flicker-banding \& moiré patterns, we introduce an ISP-based flicker-banding simulation pipeline. ISP means the image signal processor in the camera, and we utilize a pretrained Invertible-ISP~\cite{xing2021invertibleimagesignalprocessing} to approximate the transformation between RAW and sRGB domains. It models different forms of banding in the RAW domain while maintaining realistic moiré characteristics, facilitating stable and effective training.

% Our experimental results indicate the effectiveness of simulated data inclusion.

\begin{table*}[t]
\centering
% \tiny
\begin{threeparttable}
\caption{Comparison with existing screen-captured image datasets. Resolution denotes average image resolution. Artifacts means the degradation types on the screen-captured images. Scene indicates the diversity of visual content categories covered (DCID/TMM22/UHDM: natural scenes, webpages, documents; LCDMoiré: text only; RIFLE: advertisement; \datasetname: natural scenes, documents, cartoons, video games, mobile UI, webpages). Our \datasetname{} is the first dataset featuring \textbf{both artifact types}, \textbf{broader scene diversity}, and coverage of \textbf{three display technologies}. \vspace{-4pt}}
\label{tab:wide_comparison}
\resizebox{0.95\textwidth}{!}{%
\begin{minipage}{\textwidth}  
\centering
\renewcommand{\arraystretch}{1.3}
\begin{tabular}{|c|c|c|c|c|c|c|c|c|}  
\hline
\multirow{2}{*}{\textbf{Dataset}}&\multirow{2}{*}{\textbf{Size}}&\multirow{2}{*}{\textbf{Resolution}} & \multirow{2}{*}{\textbf{Scene}} & \multicolumn{2}{c|}{\textbf{Artifacts}} & \multicolumn{2}{c|}{\textbf{Devices}} & \multirow{2}{*}{\shortstack{\textbf{Real}\\\textbf{-world}}} \\
\cline{5-8}
 & & &&\textbf{Banding} & \textbf{Moiré} &\textbf{Camera} &\textbf{Display} & \\
\hline
\rowcolor{gray!5}
DCID~\cite{mei2025imagedemoireingusingdual} &8,959&$2133\times1200$& 3 & \textcolor{red}{\textbf{$\times$}} & \textcolor{green!60!black}{\textbf{$\checkmark$}} & 3 & LCD & \raisebox{-0.2ex}{\textcolor{green!60!black}{\textbf{$\checkmark$}}} \\
\hline
\rowcolor{white}
TMM22~\cite{RRID} &948&$384\times384$& 3 & \textcolor{red}{\textbf{$\times$}} & \textcolor{green!60!black}{\textbf{$\checkmark$}} & 4 & / & \raisebox{-0.2ex}{\textcolor{green!60!black}{\textbf{$\checkmark$}}} \\
\hline
\rowcolor{gray!5}
UHDM~\cite{yu2022towards} &5,000&$4328\times3248$& 3 & \textcolor{red}{\textbf{$\times$}} & \textcolor{green!60!black}{\textbf{$\checkmark$}} & 3 & LCD & \raisebox{-0.2ex}{\textcolor{green!60!black}{\textbf{$\checkmark$}}} \\
\hline
\rowcolor{white}
LCDMoiré~\cite{LCDM} &10,200&$1024\times1024$& 1 & \textcolor{red}{\textbf{$\times$}} & \textcolor{green!60!black}{\textbf{$\checkmark$}} & / & LCD& \raisebox{-0.2ex}{\textcolor{red}{\textbf{$\times$}}} \\
\hline
\rowcolor{gray!5}
RIFLE~\cite{RIFLE}&60&$4096\times3072$& 1 & \textcolor{green!60!black}{\textbf{$\checkmark$}} & \textcolor{red}{\textbf{$\times$}} & 1 & LED & \raisebox{-0.2ex}{\textcolor{green!60!black}{\textbf{$\checkmark$}}} \\
\hline
\rowcolor{blue!10}
\textbf{\textcolor{violet}{\datasetname}} &3,000& $3706\times3368$&6 & \textbf{\textcolor{green!60!black}{$\checkmark$}} & \textbf{\textcolor{green!60!black}{$\checkmark$}} & \textbf{5} & \textbf{LCD,LED,OLED} & \raisebox{-0.2ex}{\textbf{\textcolor{green!60!black}{$\checkmark$}}} \\
\hline
\end{tabular}
\end{minipage} 
} 
\end{threeparttable}
\vspace{-8pt}
\end{table*}

% At the model level, we observe that existing flicker or moiré removal methods remain insufficient even under improved data conditions. We therefore introduce a frequency-domain decomposition and re-composition module to explicitly model degradations across different frequency components, together with a trajectory alignment loss that enforces consistency in luminance variation caused by flicker-banding. These designs significantly enhance restoration quality under complex compound degradations.

As for the model design, we introduce a frequency-domain decomposition module that separates degraded inputs into three frequency components. Since both flicker-banding and moiré artifacts predominantly manifest in the mid-frequency range, we selectively suppress this band while preserving the remaining components. The resulting representation serves as the input to the downstream model, leading to substantial performance improvements despite not fully removing the artifacts. What's more, we propose a trajectory-aligned loss to regularize the denoising process by aligning feature evolution between degraded and clean inputs, stabilizing the restoration trajectory. Together, these designs enable more stable and effective joint artifact removal. 

Overall, our contributions are summarized as follows:

\begin{itemize}
    % \item We present the first \todo{refine this statement} systematic study on joint removal of moiré patterns and flicker-banding in screen-captured images, and propose a unified restoration framework.
    \item We construct a large-scale dataset containing both moiré patterns and flicker-banding, enabling effective training and evaluation for joint artifact removal.
    \item We introduce an ISP-based flicker-banding simulation pipeline that stabilizes and expands degradation distributions, largely enhancing model performance.
    \item We propose a frequency-domain segmentation module together with a trajectory-alignment loss, significantly improving restoration performance under complex moiré patterns and flicker-banding degradations.
    \item Our experimental results demonstrate that our model exhibits clear advantages in handling flicker-banding \& moiré artifacts scenarios, and shows strong practical value in real-world applications.
\end{itemize}

\section{Related Work}
\noindent\textbf{Image Demoiréing.} Early work on image demoiréing employed multi-branch architectures to address this problem~\cite{Sun_2018}. Subsequent methods have explored different signal domains: sRGB-based approaches~\cite{AFN,MMDM,Deep,AMNet}, RAW-domain processing~\cite{2025_ICML}, and hybrid sRGB+RAW pipelines~\cite{RDNet,RRID}. From a methodological perspective, multiresolution networks~\cite{2021_TPAMI,2024_Sensor}, cross-domain learning combining frequency and spatial representations~\cite{Deep,RRID}, and explicit frequency-domain transforms~\cite{2021_TPAMI,AFN,2025_TCSVT} have been widely adopted. To enhance detail recovery, multi-view fusion strategies leveraging wide-angle and ultra-wide-angle lenses~\cite{mei2025imagedemoireingusingdual} or focused and defocused image pairs~\cite{2025_TPAMI} have also been proposed. Additionally, unsupervised~\cite{2021_neuro,UnDem} and training-free methods~\cite{2020_NeurlPS} have been developed to reduce reliance on large-scale paired datasets.

\noindent\textbf{Image Debanding.} Flicker-banding manifests as periodic luminance stripes in images captured from display screens. This artifact arises from the temporal mismatch between camera acquisition and display brightness modulation~\cite{RIFLE}. Modern smartphone cameras typically employ CMOS sensors with electronic rolling shutters~\cite{durini2019high}, which expose sensor rows sequentially rather than simultaneously, introducing temporal offsets across the image. Meanwhile, displays modulate brightness through temporally varying mechanisms: OLED panels use pulse-width modulation (PWM)~\cite{geffroy2006organic}, LED displays adopt scanning refresh strategies. The interaction between these temporal dynamics produces spatially structured banding patterns. RIFLE~\cite{RIFLE} represents the only learning-based method specifically targeting flicker-banding removal. However, its performance degrades when banding coexists with other artifacts, motivating the need for unified restoration approaches.

\section{Dataset: \datasetname}
To enable systematic study of joint flicker-banding and moiré patterns removal, we construct a large-scale dataset featuring diverse degradation patterns. The data collection and capture procedures are described below.

\vspace{-2mm}
\subsection{Data Collection}
To ensure broad coverage of visual content, we aggregate source images from multiple existing datasets, as single datasets typically lack sufficient scene diversity. Specifically, we collect document images from DocBank~\cite{li2020docbank} and PubLayNet~\cite{zhong2019publaynet}, natural images from LSDIR~\cite{LSDIR}, cartoon images from Danbooru2021, video game images from VideoGameBunny~\cite{taesiri2024videogamebunnyvisionassistantsvideo}, and mobile UI screenshots from Rico~\cite{deka2017rico}. Additionally, we extract website screenshots from DSCVC~\cite{wang2025dscvc} and screen-content-dataset~\cite{scvqa2025}. In total, our dataset covers six distinct scenes.

To obtain high-quality source images, we apply IQA filtering to remove low-quality images. Images that are excessively dark or noisy are excluded, as they are less likely to exhibit pronounced banding artifacts. Additionally, images containing sensitive information are omitted to ensure privacy and compliance with ethical standards. We also adjust the proportions of images from different sources to reflect the approximate scene distribution in real-world captures.

\begin{figure*}[t]
    \centering
    \includegraphics[width=\linewidth]{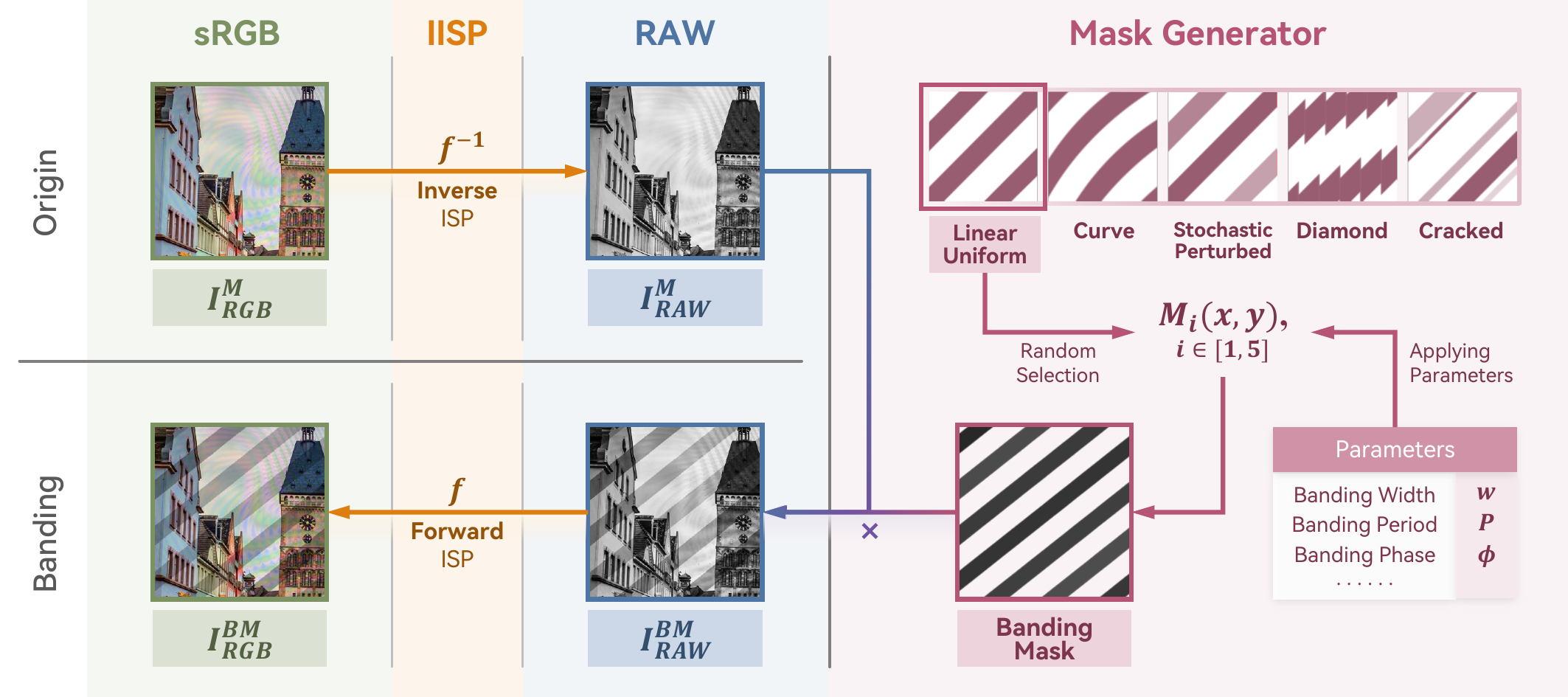}
    \caption{
        Overview of the simulation pipeline.
        Given an sRGB image with moiré, Inverse-ISP model is used to convert it to RAW format.
        Then, a flicker-banding mask is generated and applied in the RAW domain to simulate banding artifacts.
        Finally, the ISP model converts the degraded RAW image back to RGB space, producing a training pair with both moiré and banding artifacts.
    }
    \label{fig:pipeline overview}
% \vspace{-15pt}
% \vspace{-2mmsog}
\vspace{-3mm}
\end{figure*}

\subsection{Data Capture}
To accurately reflect real-world conditions, we capture degraded images by photographing display screens with smartphone cameras. To account for variations in sensor characteristics across manufacturers, we employ five different smartphone: Vivo X200s, Xiaomi 17 Pro, Huawei Mate 50, Huawei Mate 60, and iPhone 17 Pro. To cover diverse display technologies, we use LCD, LED, and OLED screens. We systematically acquire images exhibiting flicker-banding only, moiré only, and both artifacts combined (LCD excluded, as they do not produce banding artifacts).

For each source image, we display it in full-screen mode on the target display and capture it using the smartphone camera under varying exposure settings, screen brightness levels, and viewing angles. The clean ground-truth (GT) images are obtained by aligning the original source images with the captured photos using pixel-level registration techniques.

% The resulting dataset comprises approximately 3,000 image pairs in total, with 1,000 pairs for each degradation type (flicker-banding only, moiré only, and combined artifacts). We randomly partition each subset into 930 training, 20 validation, and 50 test pairs. Due to the use of different camera models, image resolutions vary (e.g., $4000\times3000$, $4096\times3072$, $3024\times4032$). A comparison with existing datasets is provided in Table~\ref{tab:wide_comparison}. The key characteristics of our dataset are summarized as follows:

The resulting dataset comprises approximately 3,000 image pairs in total, with 1,000 pairs for each degradation type (flicker-banding only, moiré only, and combined artifacts). We randomly partition each subset into 930 training, 20 validation, and 50 test pairs. A comparison with existing datasets is provided in Table~\ref{tab:wide_comparison}. The key characteristics of our dataset are summarized as follows:

\vspace{-2mm}
\begin{itemize}
    \item \textbf{Diverse scene content.} The dataset spans six scene categories: natural images, websites, documents, cartoons, video games, and mobile UI.
    \item \textbf{Comprehensive artifact coverage.} The dataset includes images with banding, moiré, and both artifacts combined, captured under varying viewing angles and distances to ensure diverse artifact morphologies.
    \item \textbf{Real-world capture conditions.} All images are captured using consumer smartphone cameras of different kinds on commercial LCD, LED, and OLED displays, reflecting practical usage scenarios.
\end{itemize}

\section{Methods}
Based on our backbone, PiSA-SR~\cite{sun2024pisasr}, we introduce following components to enhance the model performance in removing the flicker-banding and moiré.

\begin{figure}[t]
    \centering
    \includegraphics[width=\columnwidth]{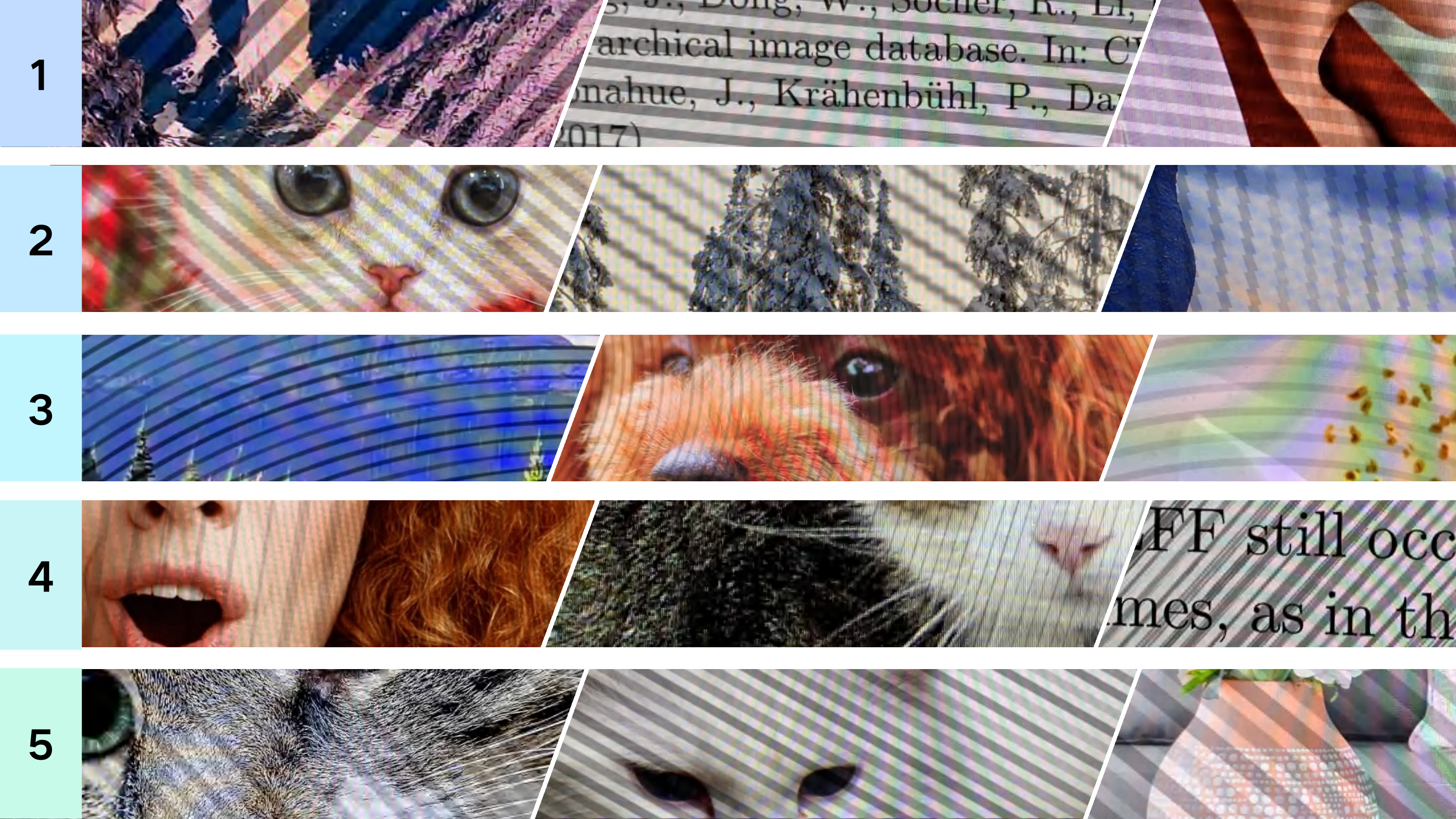}
    % \vspace{-6mm}
    \caption{Example of simulation banding types. 1: simple; 2: diamond; 3: curve; 4: cracked; 5: complex. More examples are in the supplementary material.}
    \label{fig:example of simulation}
% \vspace{-20pt}
% \vspace{-7mm}
\end{figure}

\subsection{ISP-based Simulation Pipeline (ISPS)}
% \vspace{-2mm}
While real-world data provides diverse compound degradations, we observe that certain banding patterns occur infrequently that hinder robust model training. Moreover, directly training on real data with highly entangled artifacts often results in unstable convergence. So we propose a physically-motivated RAW-domain simulation pipeline that synthesizes flicker-banding on moiré images, enabling controlled augmentation of underrepresented degradation types while preserving realistic artifact characteristics.

% \vspace{-1mm}
\noindent\textbf{RAW-Domain Simulation.} A key insight of our approach is that flicker-banding originates from temporal luminance modulation during sensor exposure, which is more accurately modeled in the linear RAW domain rather than the nonlinear sRGB space.

Therefore, we adopt the Inverse-ISP model $f^{-1}$~\cite{xing2021invertibleimagesignalprocessing} to transform moiré-corrupted sRGB images $I_{sRGB}^M$ from the UHDM dataset~\cite{yu2022towards} into pseudo-RAW representations: 
\vspace{-3mm}
\begin{equation}
    I_{RAW}^M=f^{-1}(I_{sRGB}^M).
\end{equation}
This enables physically plausible banding synthesis that respects the camera imaging pipeline.

\noindent\textbf{Comprehensive Banding Pattern Coverage.} Real-world flicker-banding exhibits diverse morphologies depending on the interaction between camera, display and shooting conditions. To ensure sufficient coverage, we design a parameterized mask generator $M(x,y)$ that produces five representative banding types, ranging from regular parallel stripes to curved, grid-like, fragmented, and stochastically perturbed structures (see Figure~\ref{fig:example of simulation}), which could be further controlled by parameters including stripe width, period, phase, and feathering radius. This process effectively covers the vast majority of banding patterns encountered in everyday screen photography. Detailed mathematical formulations of each pattern type are provided in the supplementary material.

\noindent\textbf{Gain-Based Degradation Model.} The generated mask is converted to spatially-varying gain coefficients that model light intensity attenuation during exposure. We incorporate a baseline darkness parameter $D$ and per-stripe brightness jitter $\eta_k$ to simulate the stochastic nature of real banding:
\begin{equation}
    G(x,y) = \max\left( G_{min}, \; 1 - M(x,y) \cdot (1 - D) \cdot \eta_k \right),
\end{equation}
where $G_{min}$ prevents complete pixel extinction. The banded RAW image is obtained via:
\begin{equation}
    I_{RAW}^{BM}(x,y) = I_{RAW}^M(x,y) \cdot G(x,y).
\end{equation}
The final composite image is produced by the forward ISP: $I_{sRGB}^{BM}=f(I_{RAW}^{BM})$. This simulation expands the effective training distribution, enabling the model to generalize to rare but practically important banding patterns.

\begin{figure*}[t]
    \centering
    \includegraphics[width=\textwidth]{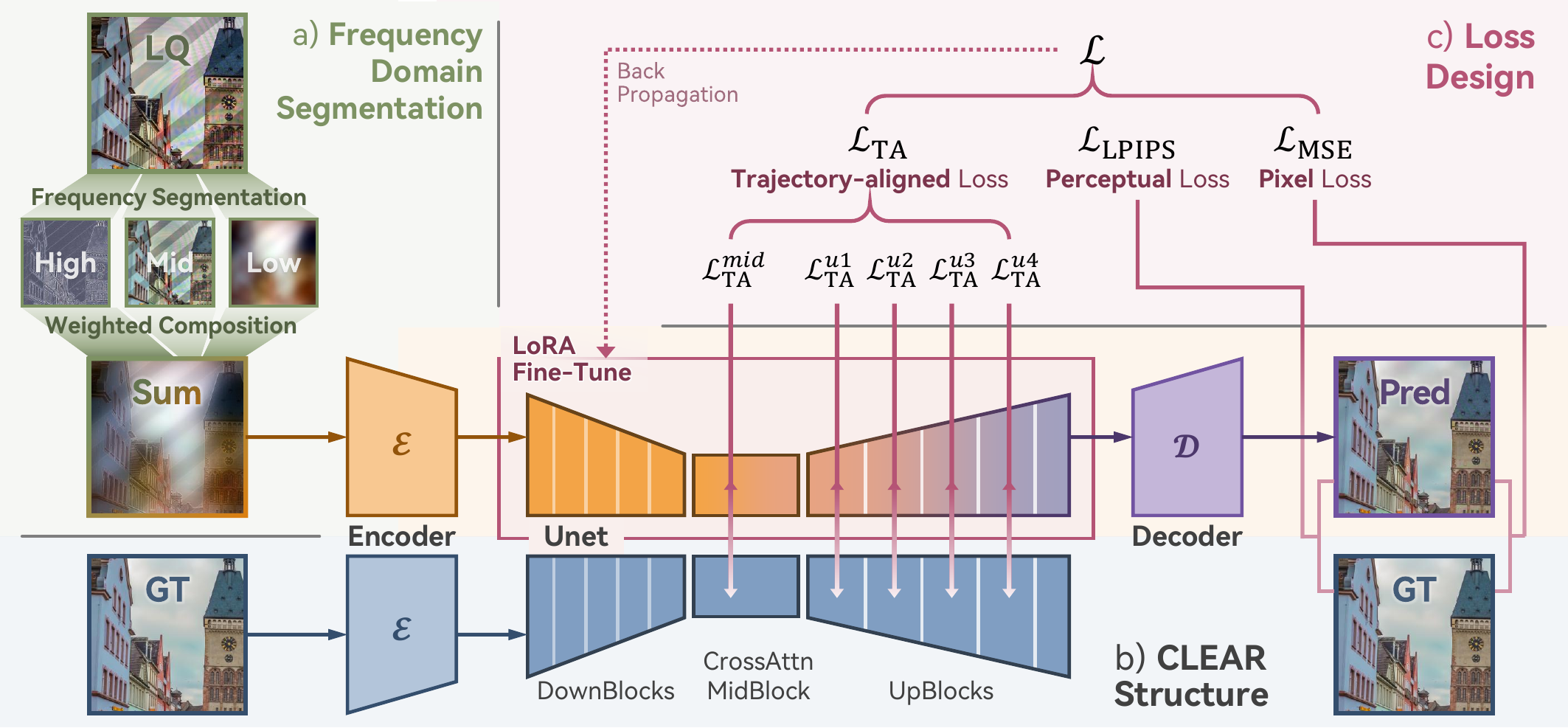}
    \vspace{-5mm}
    \caption{
        Overview of the proposed CLEAR framework.
        \textbf{(a)} Frequency-domain segmentation (FS) module.
        \textbf{(b)} The overall CLEAR architecture.
        \textbf{(c)} The training objective combines trajectory alignment loss (TA) with perceptual and pixel losses.
    }
    \vspace{-5mm}
    \label{fig:structure}
\end{figure*}

\subsection{Frequency Domain Segmentation (FS)}

Moiré patterns and flicker-banding exhibit distinct spectral characteristics: moiré artifacts are associated with periodic or quasi-periodic mid- to high-frequency interference, while flicker-banding manifests as low- to mid-frequency luminance modulation with directional structure. When both degradations coexist, directly modeling them in the spatial domain forces the network to simultaneously suppress high-frequency texture corruption and restore global luminance consistency, which often leads to optimization conflicts and unstable training. Motivated by this observation, we introduce a frequency-domain decomposition module that enables explicit manipulation of different spectral components. It allows the network to handle each frequency band in a more targeted manner as a plug-and-play component.

Given input $\mathbf{x}\in\mathbb{R}^{B\times C\times H\times W}$, we compute the Fourier transform $\mathbf{X} = \mathcal{F}(\mathbf{x})$ and define radial distance $d(u,v)$ on a normalized frequency plane with aspect-ratio correction. To partition the spectrum into frequency bands, we adopt $n$-th order Butterworth low-pass masks:
\begin{equation}
\label{eq:butterworth_lp}
B_{\mathrm{lp}}(d;\rho,n) = \frac{1}{1 + \left({d}/({\rho + \varepsilon})\right)^{2n}},
\end{equation}
which provide smooth and differentiable transitions, effectively avoiding ringing artifacts caused by hard cutoffs. 

We denotie the mask matrices as $\mathbf{B}_{b} \in \mathbb{R}^{H \times W}$ for $b \in \{\text{low}, \text{mid}, \text{high}\}$.
Using two cutoff ratios $\rho_1 < \rho_2$, we construct three complementary band masks:
\begin{equation}
\label{eq:three_band_masks}
\begin{cases}
\mathbf{B}_{\text{low}}(u,v) &= B_{\mathrm{lp}}(d(u,v);\rho_1,n), \\
\mathbf{B}_{\text{mid}}(u,v) &= B_{\mathrm{lp}}(d(u,v);\rho_2,n) - B_{\text{low}}(u,v), \\
\mathbf{B}_{\text{high}}(u,v) &= 1 - B_{\mathrm{lp}}(d(u,v);\rho_2,n),
\end{cases}
\end{equation}
where $(u,v)$ indexes the frequency plane. These masks satisfy $B_{\text{low}} + B_{\text{mid}} + B_{\text{high}} \approx 1$ pointwise. To maintain consistent partitioning across varying image sizes, we apply resolution-adaptive scaling: $\tilde{\rho}_i = \rho_i \cdot ({m_0}/{m})^{1/2}$, where $m=\min(H,W)$ and $m_0$ is a reference size.

Exploiting Fourier linearity, we perform weighted recomposition directly in frequency domain and transform it back to the pixel domain. The reconstructed image is expressed as:
\begin{equation}
\label{eq:freq_splicing}
\mathbf{I}_{\mathrm{sum}} = \Re\left\{ \mathcal{F}^{-1}\big(\mathbf{F}_{\mathrm{sum}}\big) \right\},
\end{equation}
where the composite spectrum is
\begin{equation}
\label{eq:freq_sum}
\mathbf{F}_{\mathrm{sum}} = \sum_{b} w_{b} \cdot (\mathbf{X} \odot \mathbf{B}_{b}),
\vspace{-2mm}
\end{equation}
with weights $\{w_{\text{low}}, w_{\text{mid}}, w_{\text{high}}\}$ that allow the model to balance contributions from each band. 

% From a modeling perspective, the proposed module introduces an explicit spectral inductive bias: high-frequency bands primarily capture moiré-related periodic interference, while low- and mid-frequency bands correspond to flicker-induced luminance modulation. This disentangled representation facilitates more stable training and effective recovery under compound degradations.

This module serves as a frequency-aware front-end that reshapes the degraded input into a more structured and learnable representation. Through weighted recomposition in the frequency domain followed by inverse transformation, it selectively suppresses frequency ranges that are most affected by the degradation (\eg, moiré and flicker-banding) while retaining stable structural and detail information. In addition, resolution-adaptive cutoff scaling maintains consistent frequency partitioning across different image sizes, improving robustness under varying resolutions.

Although this module alone does not fully eliminate artifacts, it provides partial suppression of flicker-banding and moiré artifacts, and provides a better-conditioned input for subsequent networks, leading to more stable optimization and improved restoration performance.

\subsection{Trajectory-aligned Loss (TA)}
When moiré patterns and flicker-banding coexist, the restoration process must not only recover spatial details but also follow a \emph{consistent correction trajectory} across the network hierarchy and diffusion steps. flicker-banding perturbs luminance-related components and induces structured intensity modulation, which can cause the network to drift toward unstable intermediate representations during denoising. This drift is amplified by high-frequency moiré interference, making the optimization landscape ill-conditioned when training with pixel-level losses alone. To address this, we introduce a \emph{trajectory alignment loss} that constrains feature evolution from low-quality (LQ) inputs to match that from ground-truth (GT) targets at the \emph{same diffusion timestep}, thereby stabilizing the denoising trajectory.

Let $\mathcal{U}_{\theta}$ denote the diffusion U-Net. Given paired inputs $(\mathbf{x}_{\mathrm{lq}}, \mathbf{x}_{\mathrm{gt}})$ at timestep $t$, we extract features $\mathbf{F}^{\ell}_{\mathrm{lq}}(t), \mathbf{F}^{\ell}_{\mathrm{gt}}(t) \in \mathbb{R}^{N \times C \times H \times W}$ from target layers $\ell \in \mathcal{S}$. To align feature \emph{trajectories} rather than raw magnitudes, we adopt per-channel cosine similarity, which removes scale ambiguity and focuses on directional consistency. 

For each sample $n$ and channel $c$, we vectorize the spatial dimensions and $\ell_2$-normalize:
\vspace{-1mm}
\begin{equation}
\label{eq:repa_norm}
\hat{\mathbf{v}}^{\ell}(n,c,t) = \frac{\mathrm{vec}(\mathbf{F}^{\ell}(t))_{n,c,:}}{\|\mathrm{vec}(\mathbf{F}^{\ell}(t))_{n,c,:}\|_2 + \varepsilon},
\vspace{-2mm}
\end{equation}
where $\varepsilon$ ensures numerical stability. The layer-wise trajectory alignment loss is then defined as the mean cosine distance between LQ and GT representations:
\begin{equation}
\label{eq:repa_layer_loss}
\mathcal{L}^{\ell}_{\mathrm{TA}}(t)
= \frac{1}{N C}
\sum_{n,c}
\big( 1 - \langle \hat{\mathbf{v}}^{\ell}_{\mathrm{lq}}(n,c,t), \hat{\mathbf{v}}^{\ell}_{\mathrm{gt}}(n,c,t) \rangle \big),
\end{equation}
where $\langle \mathbf{v}_1,\mathbf{v}_2 \rangle$ means cosine similarity between $\mathbf{v}_1$ and $\mathbf{v}_2$.

The trajectory-aligned loss aggregates across selected layers with weights $\{\lambda_{\ell}\}_{\ell \in \mathcal{S}}$ and a global factor $\gamma$:
\begin{equation}
\label{eq:repa_total_loss}
\mathcal{L}_{\mathrm{TA}}(t)
= \gamma \sum_{\ell \in \mathcal{S}}
\lambda_{\ell} \, \mathcal{L}^{\ell}_{\mathrm{TA}}(t).
\vspace{-2mm}
\end{equation}
By enforcing per-channel directional agreement in representation space, the trajectory-aligned loss focuses on \emph{structural consistency of feature trajectories} rather than absolute activation magnitudes, which is particularly effective for stabilizing luminance correction while preserving the high-frequency details for moiré suppression.

% \noindent\textbf{Overall Training Objective.} The complete training objective combines the trajectory alignment loss with standard reconstruction losses in a weighted manner:
\subsection{Overall Training Objective.}
The complete training objective combines the trajectory-aligned loss with standard reconstruction losses:
\begin{equation}
\label{eq:total_loss}
\mathcal{L}_{\mathrm{total}} = \lambda_{\mathrm{TA}} \mathcal{L}_{\mathrm{TA}} + \lambda_{\mathrm{LPIPS}} \mathcal{L}_{\mathrm{LPIPS}} + \lambda_{\mathrm{MSE}} \mathcal{L}_{\mathrm{MSE}},
\end{equation}
where $\mathcal{L}_{\mathrm{MSE}}$ is the mean squared error (MSE) loss for pixel-level reconstruction, and $\mathcal{L}_{\mathrm{LPIPS}}$ is the LPIPS~\cite{zhang2018unreasonable} perceptual loss that measures feature-level similarity using pretrained image evaluation networks.

We adopt a parameter-efficient fine-tuning strategy using LoRA~\cite{hu2022lora}, which inserts low-rank adaptation matrices into the diffusion U-Net. During training, only the LoRA parameters in the U-Net blocks are optimized, while the encoder and decoder remain frozen. This design preserves the pretrained representation capacity while enabling efficient adaptation to the joint artifact removal task. Together, these complementary objectives enable effective joint removal of flicker-banding and moiré artifacts.

\begin{table*}[h]
\centering
\scriptsize
\caption{Quantitative results of different methods on \textbf{cropped} \datasetname{} testing dataset. Compared methods are retrained with \datasetname's training dataset. The best and second best results are colored with \textbf{\textcolor{red}{red}} and \textbf{\textcolor{blue}{blue}}. CLEAR gains a significant advantage over other methods.}
\vspace{-2mm}

\setlength{\tabcolsep}{4pt}
\renewcommand{\arraystretch}{1.15}

\begin{tabularx}{\textwidth}{
lc
*{6}{>{\centering\arraybackslash}X}
}
\toprule
\textbf{Method}
& Task
& SSIM $\uparrow$
& ms-SSIM $\uparrow$
& LPIPS $\downarrow$
& DISTS $\downarrow$
& FSIM $\uparrow$
& GMSD $\downarrow$ \\
\midrule
\rowcolor{gray!5}
LQ & -
& 0.5244 & 0.6495 & 0.5210 & 0.3003 & 0.7437 & 0.2064 \\
\rowcolor{white}
NeRD-Rain~\cite{NeRD-Rain} & Deraining
& 0.6685 & 0.7254 & 0.5446 & 0.3130 & 0.7801 & 0.1723 \\
\rowcolor{gray!5}
ESDNet~\cite{yu2022towards} & Demoiré
& \textcolor{red}{\textbf{0.7354}} & \textcolor{blue}{\textbf{0.7704}} & 0.4983 & 0.3107 & \textcolor{blue}{\textbf{0.8030}} & 0.1707 \\
\rowcolor{white}
MAT~\cite{xie2025mat} & Super-resolution
& 0.6732 & 0.7203 & 0.5346 & 0.3095 & 0.7835 & 0.1648 \\
\rowcolor{gray!5}
ResShift~\cite{yue2023resshift} & Super-resolution
& 0.7099 & 0.7404 & 0.4548 & \textcolor{blue}{\textbf{0.2709}} & 0.7912 & 0.1812 \\
\rowcolor{white}
PiSA-SR~\cite{sun2024pisasr}  & Super-resolution
& 0.6598 & 0.7236 & \textcolor{blue}{\textbf{0.4191}} & 0.2738 & 0.7646 & 0.2045 \\
\rowcolor{gray!5}
InvSR~\cite{yue2025InvSR}  & Super-resolution
& 0.7037 & 0.6988 & 0.4391 & 0.3164 & 0.7576 & 0.2011 \\
\rowcolor{white}
RIFLE~\cite{RIFLE}  & Debanding
&0.6391	&0.7470	&0.4557	&0.2790	&0.7983	&\textcolor{blue}{\textbf{0.1631}}\\
\rowcolor{blue!10}
\textbf{\textcolor{violet}{CLEAR}}
& Debanding \& Demoire
&  \textcolor{blue}{\textbf{0.7163}}
& \textcolor{red}{\textbf{0.8159}}
& \textcolor{red}{\textbf{0.3183}}
& \textcolor{red}{\textbf{0.2354}}
& \textcolor{red}{\textbf{0.8249}}
& \textcolor{red}{\textbf{0.1598}} \\

\bottomrule
\end{tabularx}
\vspace{-1mm}
\label{tab:quantitative_results_bm3k}
\end{table*}

\begin{table*}[t]
\centering
\vspace{-2mm}

\begin{minipage}[t]{0.35\linewidth}
\centering
\captionof{table}{Comparison with combined models.}
\vspace{-2mm}
\scalebox{0.95}{
\scriptsize
\begin{tabular}{c|ccc}
\toprule
Metrics & LQ & RIFLE \& ESDNet & CLEAR \\
\midrule
ms-SSIM$\uparrow$ & 0.6495 & 0.7395 & \textcolor{red}{\textbf{0.8159}} \\
LPIPS$\downarrow$   & 0.5210 & 0.4438 & \textcolor{red}{\textbf{0.3183}} \\
DISTS$\downarrow$ & 0.3003 & 0.2831 & \textcolor{red}{\textbf{0.2354}} \\
\bottomrule
\end{tabular}
}
\label{tab:combined model table}
\end{minipage}
\hfill
\begin{minipage}[t]{0.35\linewidth}
\centering
\captionof{table}{Ablation on simulation pipeline.}
\vspace{-2mm}
\scalebox{0.95}{
\scriptsize
\begin{tabular}{c|ccc}
\toprule
Metrics &  \datasetname & ISPS & \datasetname+ISPS  \\
\midrule
ms-SSIM$\uparrow$   & 0.7236 & 0.7974 & \textcolor{red}{\textbf{0.8023}}  \\
LPIPS$\downarrow$  & 0.4191 & 0.3920 & \textcolor{red}{\textbf{0.3897}}  \\
DISTS$\downarrow$  & 0.2738 & 0.2611 & \textcolor{red}{\textbf{0.2592}}  \\
\bottomrule
\end{tabular}
}
\label{tab:simulation ablation}
\end{minipage}
\hfill
\begin{minipage}[t]{0.29\linewidth}
\centering
\captionof{table}{Ablation on model design.}
\vspace{-2mm}
\scalebox{0.95}{
\scriptsize
\begin{tabular}{c|ccc}
\toprule
Metrics & Baseline & FS & TA + FS \\
\midrule
ms-SSIM$\uparrow$   & 0.8023 & 0.8093 & \textcolor{red}{\textbf{0.8159}} \\
LPIPS$\downarrow$     & 0.3897 & 0.3610 & \textcolor{red}{\textbf{0.3183}} \\
DISTS$\downarrow$   & 0.2592 & 0.2489 & \textcolor{red}{\textbf{0.2354}} \\
\bottomrule
\end{tabular}
}
\label{tab:model ablation}
\end{minipage}
\vspace{-6mm}
\end{table*}

% \begin{figure}[t]
% \centering
% % ===================== Row 1 =====================
% \begin{minipage}[t]{0.235\textwidth}
%     \centering
%     \includegraphics[width=\linewidth]{figures/visual/selected_images_RIFLE_ESDNet/GT/000067.png} \\
%     \footnotesize GT
% \end{minipage}
% \hfill
% \begin{minipage}[t]{0.235\textwidth}
%     \centering
%     \includegraphics[width=\linewidth]{figures/visual/selected_images_RIFLE_ESDNet/LQ/000067.png} \\
%     \footnotesize LQ
% \end{minipage}
% \vspace{1mm}
% \begin{minipage}[t]{0.235\textwidth}
%     \centering
%     \includegraphics[width=\linewidth]{figures/visual/selected_images_RIFLE_ESDNet/RIFLE_ESDNet/000067.jpg}\\
%     \footnotesize RIFLE \& ESDNet
% \end{minipage}
% \hfill
% \begin{minipage}[t]{0.235\textwidth}
%     \centering
%     \includegraphics[width=\linewidth]{figures/visual/selected_images_RIFLE_ESDNet/CLEAR/000067.png}\\
%     \footnotesize CLEAR
% \end{minipage}
% \vspace{-2.5mm}
% \caption{Visual comparison with combined methods. }
% \label{fig:Visual comparison of comblined models}
% \vspace{-7mm}
% \end{figure}

\begin{figure}[t]
\centering
% ===================== Row 1 =====================
\begin{minipage}[t]{0.155\textwidth}
    \centering
    \croppedimg{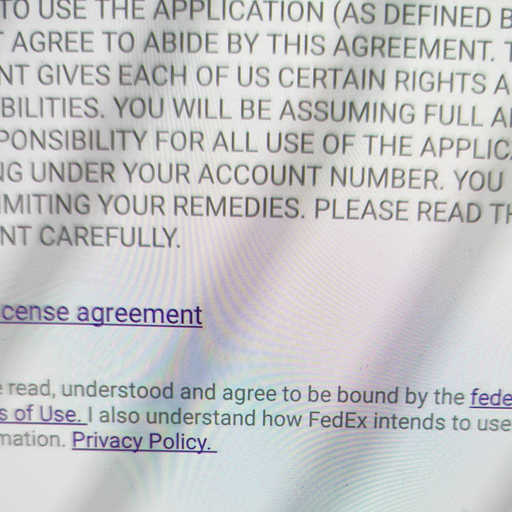} \\
    \footnotesize LQ
\end{minipage}
\vspace{1mm}
\begin{minipage}[t]{0.155\textwidth}
    \centering
    \croppedimg{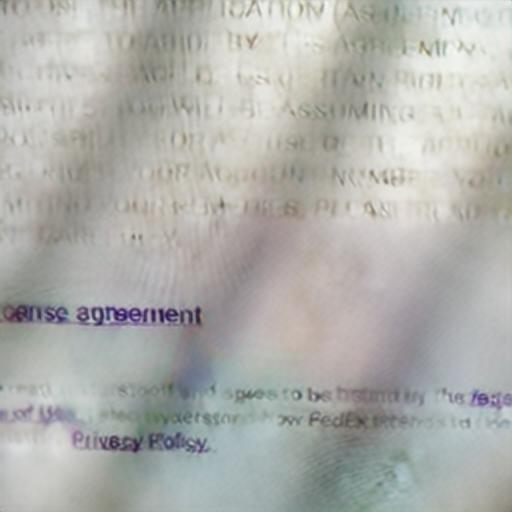}\\
    \footnotesize RIFLE \& ESDNet
\end{minipage}
\hfill
\begin{minipage}[t]{0.155\textwidth}
    \centering
    \croppedimg{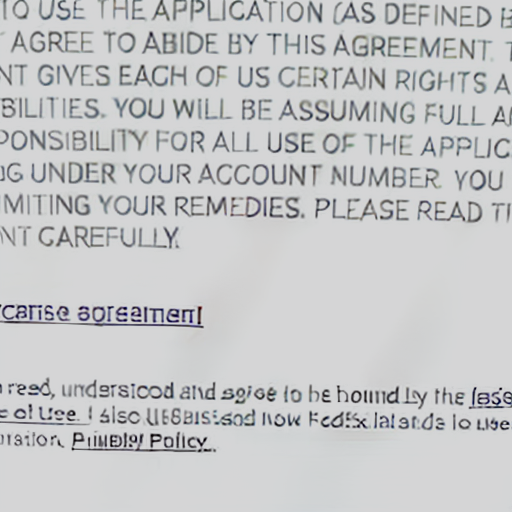}\\
    \footnotesize CLEAR
\end{minipage}
\vspace{-2mm}
\caption{Visual comparison with combined methods. }
\label{fig:Visual comparison of comblined models}
\vspace{-5mm}
\end{figure}

\begin{figure*}[t]
\centering

% ===================== Row 1 =====================
\begin{minipage}[t]{0.19\textwidth}
    \centering
    \croppedimg{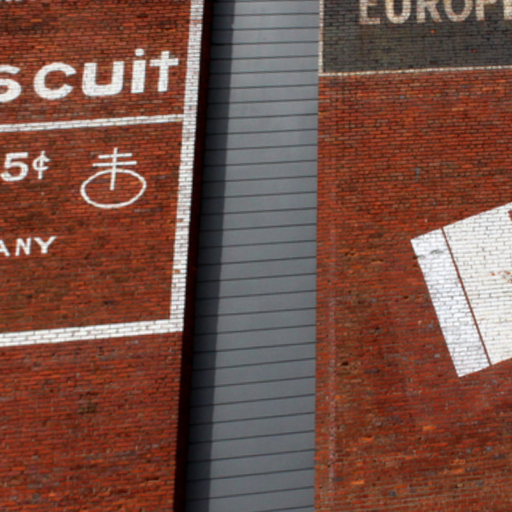}\\
    \footnotesize GT
\end{minipage}
\hfill
\begin{minipage}[t]{0.19\textwidth}
    \centering
    \croppedimg{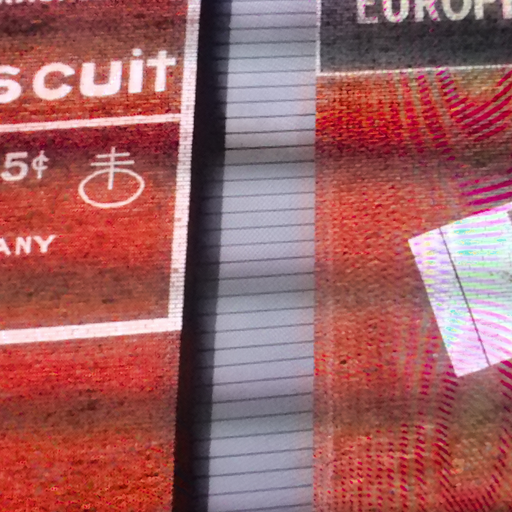}\\
    \footnotesize LQ
\end{minipage}
\hfill
\begin{minipage}[t]{0.19\textwidth}
    \centering
    \croppedimg{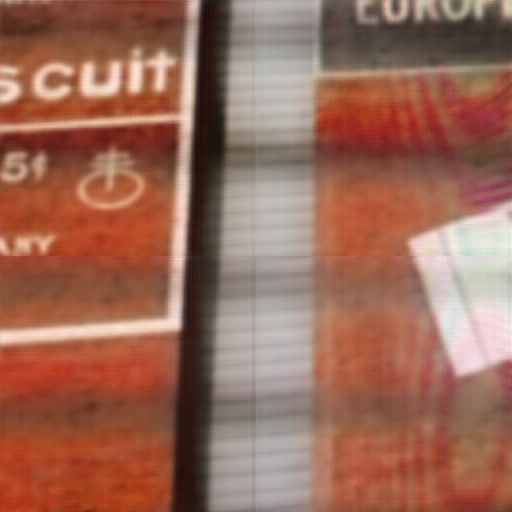}\\
    \footnotesize NeRD-Rain
\end{minipage}
\hfill
\begin{minipage}[t]{0.19\textwidth}
    \centering
    \croppedimg{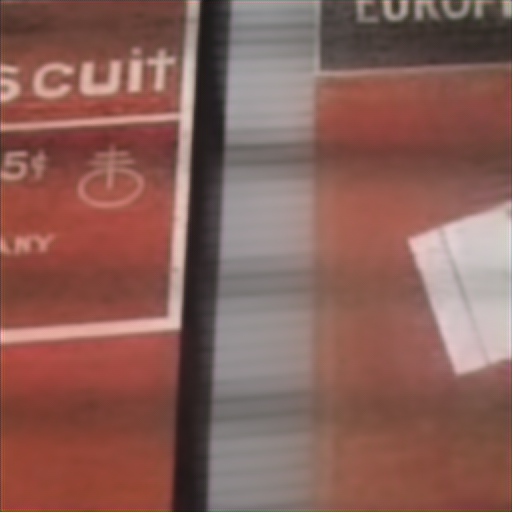}\\
    \footnotesize ESDNet
\end{minipage}
\hfill
\begin{minipage}[t]{0.19\textwidth}
    \centering
    \croppedimg{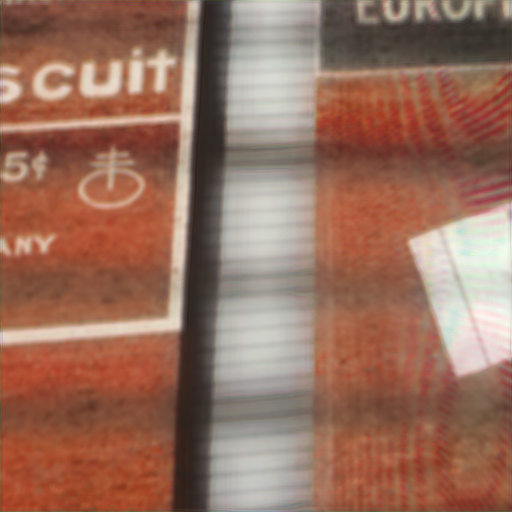}\\
    \footnotesize MAT
\end{minipage}
\vspace{1mm}
% ===================== Row 2 =====================
\begin{minipage}[t]{0.19\textwidth}
    \centering
    \croppedimg{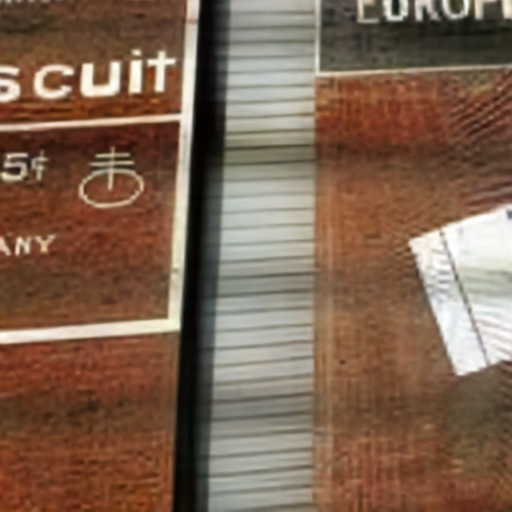}\\
    \footnotesize ResShift
\end{minipage}
\hfill
\begin{minipage}[t]{0.19\textwidth}
    \centering
    \croppedimg{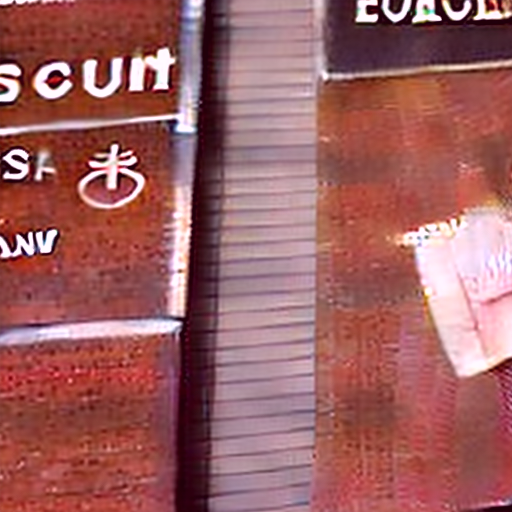}\\
    \footnotesize PiSA-SR
\end{minipage}
\hfill
\begin{minipage}[t]{0.19\textwidth}
    \centering
    \croppedimg{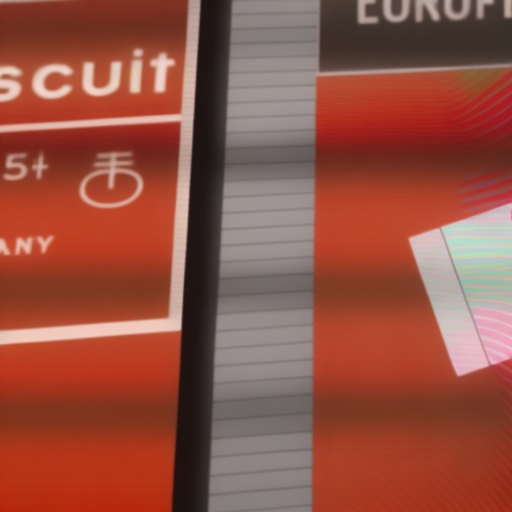}\\
    \footnotesize InvSR
\end{minipage}
\hfill
\begin{minipage}[t]{0.19\textwidth}
    \centering
    \croppedimg{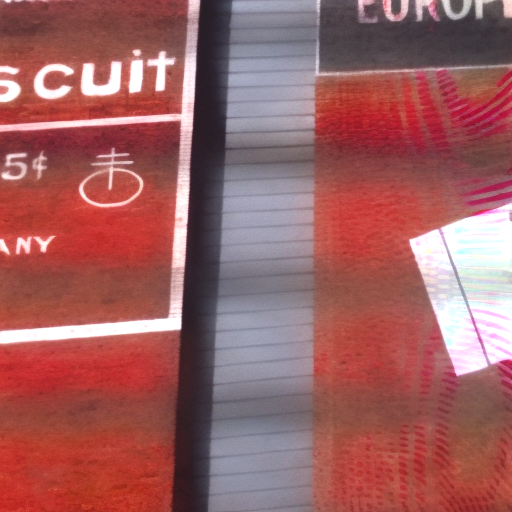}\\
    \footnotesize RIFLE
\end{minipage}
\hfill
\begin{minipage}[t]{0.19\textwidth}
    \centering
    \croppedimg{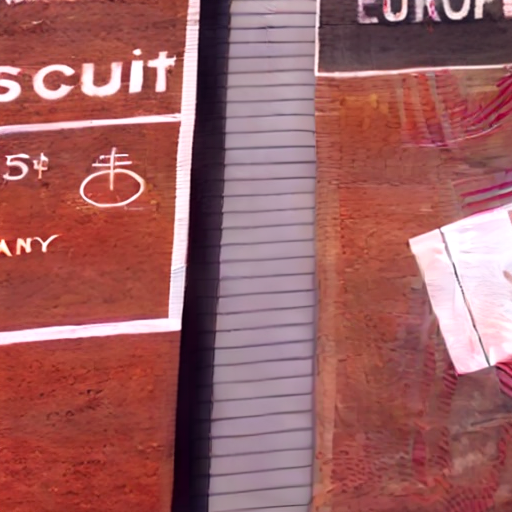}\\
    \footnotesize CLEAR
\end{minipage}

\vspace{0.5mm}
{\color{gray}\rule{\textwidth}{0.6pt}}
\vspace{-2.5mm}

\begin{minipage}[t]{0.19\textwidth}
    \centering
    \croppedimg{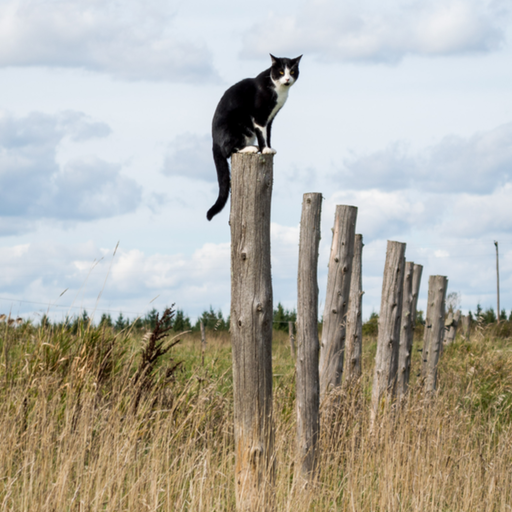}\\
    \footnotesize GT
\end{minipage}
\hfill
\begin{minipage}[t]{0.19\textwidth}
    \centering
    \croppedimg{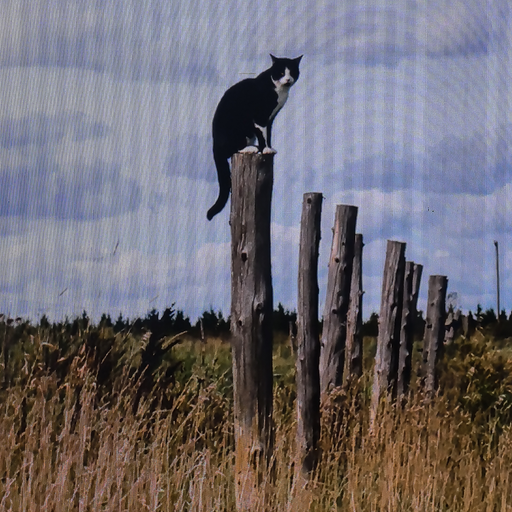}\\
    \footnotesize LQ
\end{minipage}
\hfill
\begin{minipage}[t]{0.19\textwidth}
    \centering
    \croppedimg{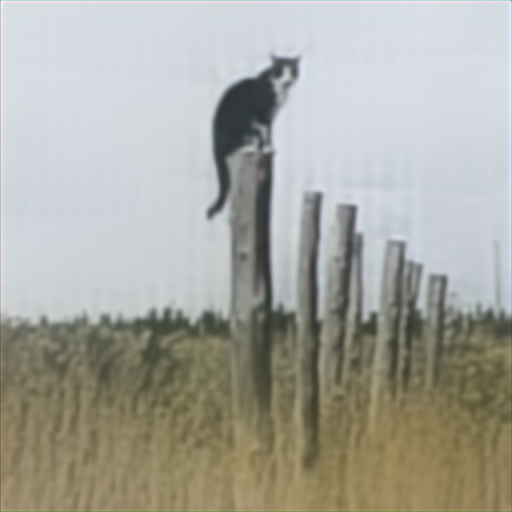}\\
    \footnotesize ESDNet
\end{minipage}
\hfill
\begin{minipage}[t]{0.19\textwidth}
    \centering
    \croppedimg{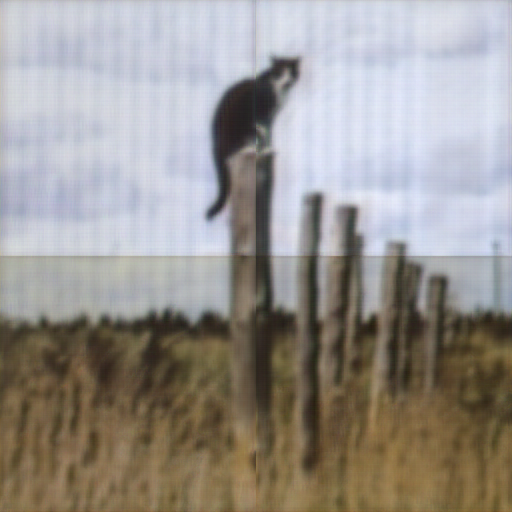}\\
    \footnotesize NeRD-Rain
\end{minipage}
\hfill
\begin{minipage}[t]{0.19\textwidth}
    \centering
    \croppedimg{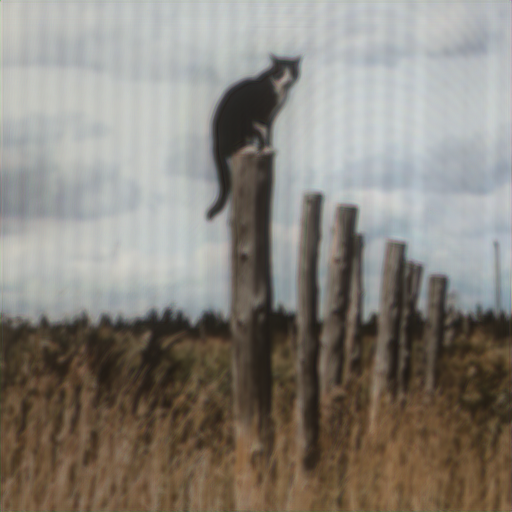}\\
    \footnotesize MAT
\end{minipage}
% \vspace{-2mm}
% ===================== Row 2 =====================
\begin{minipage}[t]{0.19\textwidth}
    \centering
    \croppedimg{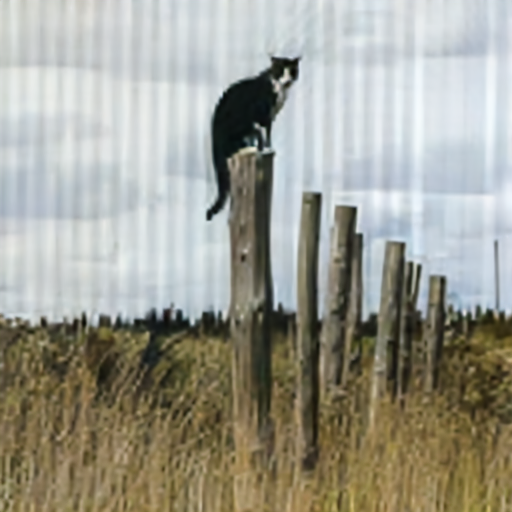}\\
    \footnotesize ResShift
\end{minipage}
\hfill
\begin{minipage}[t]{0.19\textwidth}
    \centering
    \croppedimg{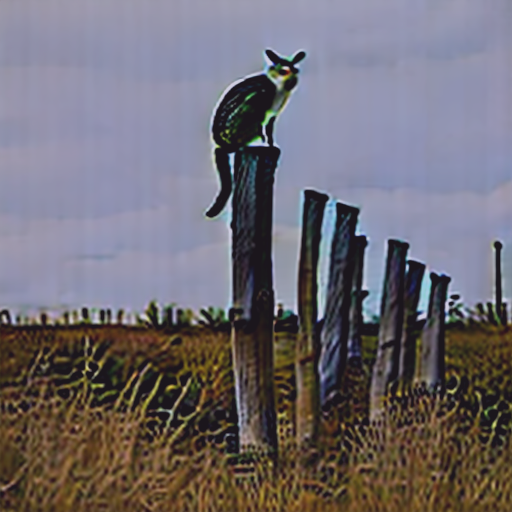}\\
    \footnotesize PiSA-SR
\end{minipage}
\hfill
\begin{minipage}[t]{0.19\textwidth}
    \centering
    \croppedimg{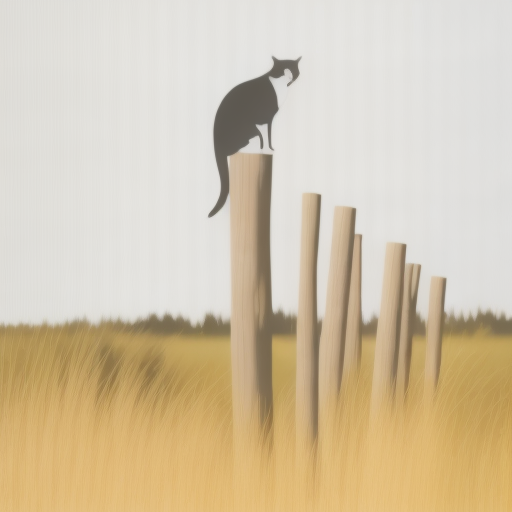}\\
    \footnotesize InvSR
\end{minipage}
\hfill
\begin{minipage}[t]{0.19\textwidth}
    \centering
    \croppedimg{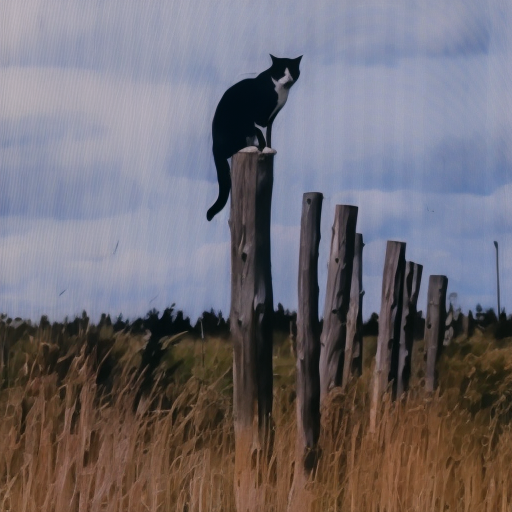}\\
    \footnotesize RIFLE
\end{minipage}
\hfill
\begin{minipage}[t]{0.19\textwidth}
    \centering
    \croppedimg{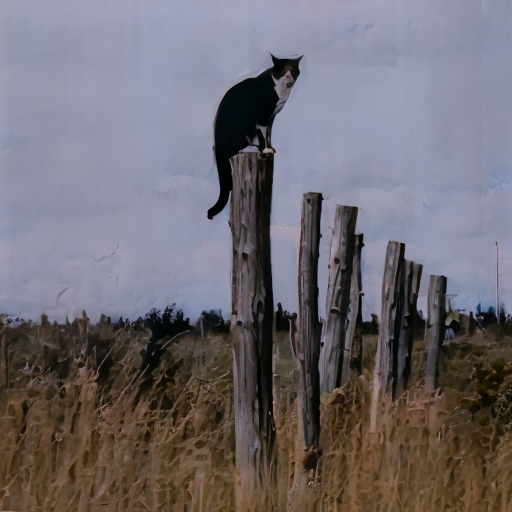}\\
    \footnotesize CLEAR
\end{minipage}
\caption{Visual comparison with flicker-banding \& moiré images (LQ), clean images (GT), and other compared methods on the \datasetname{} testing dataset. Compared methods are retrained with \datasetname's training dataset. CLEAR gains great advantages over other methods.}
\label{fig:main visual comparison}
\vspace{-2mm}
\end{figure*}

\newpage

\section{Experiments}
\vspace{-1mm}
\subsection{Experiments Setup}
\vspace{-1mm}
\noindent\textbf{Data Construction.} The dataset we mainly used for training and testing is our proposed dataset, \datasetname, consisting of the scenes where the flicker-banding and moiré coexist. For our proposed simulation pipeline, we use UHDM~\cite{yu2022towards} for introducing the real-world moiré artifacts. It is worth noting that we crop the \datasetname{} testing dataset to 512$\times$512 to highlight the evaluation metrics difference.

% \vspace{-0.5mm}
\noindent\textbf{Evaluation Metrics.}  Reference-based evaluation metrics are mainly used for evaluation, including SSIM~\cite{wang2004image}, ms-SSIM~\cite{wang2003multiscale}, LPIPS~\cite{zhang2018unreasonable}, DISTS~\cite{ding2020image}, FSIM~\cite{zhang2011fsim}, and GMSD~\cite{Xue2014Gradient}. As mentioned in RIFLE~\cite{RIFLE}, non-reference evaluation metrics are not suitable for evaluation on flicker-banding.

\noindent\textbf{Implementation Details.} We set the rank to 32 with a learning rate of $5\times10^{-5}$ for LoRA finetuning. The training process is performed using images of resolution $512\times512$. 

% \vspace{-2mm}
\noindent\textbf{Compared Methods.} We select the compared methods in other image reconstruction tasks (\eg, Deraining, Demoiré, Debanding, and Super-resolution), due to lack of research on removing the flicker-banding and moiré simultaneously.  We retrain the selected compared methods with our proposed dataset to ensure a fair comparison. The selected models also cover different model structures, including transformer, multi-step diffusion, and one-step diffusion models, which are always employed in image restoration tasks.

\newpage

\vspace{-3.5mm}
\subsection{Comparison with combined models}
\vspace{-2mm}
To demonstrate the necessity and value of our model, we construct a combined baseline for comparison. Specifically, we cascade the current state-of-the-art debanding model RIFLE~\cite{RIFLE} and demoiré ESDNet~\cite{yu2022towards}, evaluating this combined model on our testing dataset. The quantitative results and visual comparison with our method are presented in Tab.~\ref{tab:combined model table} and Fig.~\ref{fig:Visual comparison of comblined models}. The experimental results show that the performance of the combined model is far from satisfactory, indicating that simply integrating a strong deflickering model with a strong demoiré model is insufficient to handle scenarios where both degradations coexist. It demonstrates that our work on removing flicker-banding \& moiré is of clear practical significance.

\vspace{-3mm}
\subsection{Main Results}
\vspace{-1.8mm}
\noindent\textbf{Quantitative Results.} We provide the quantitative results of different methods on cropped \datasetname{} testing dataset in Tab.~\ref{tab:quantitative_results_bm3k}. It is obvious that \textbf{CLEAR} consistently outperforms existing methods across most evaluation metrics in the removal of flicker-banding \& moiré. Although ESDNet gains little advantage in SSIM, CLEAR performs better across a wider range of metrics, demonstrating superior model performance. What's more, we observe that even when the model achieves a substantial improvement in removing flicker-banding artifacts, the corresponding gains in quantitative evaluation metrics remain limited. Existing evaluation metrics mainly measure the overall discrepancy between the restored images and the clean ground truth images, but they are not sensitive to the flicker-banding artifacts.

\vspace{-1.5mm}
\noindent\textbf{Visual Comparison.} The visual performance comparison results are presented in Figs.~\ref{fig:main visual comparison}. CLEAR demonstrates a clear advantage in visual quality. It effectively removes severe flicker-banding while simultaneously suppressing moiré patterns, even in challenging regions with dense stripes and high-frequency content. The restored results of the compared methods exhibit various issues: severely blurred, significant noise and structural degradation, or noticeable residual flicker-banding and moiré artifacts.

\newpage

\subsection{Ablation Study}

\noindent\textbf{Simulation Pipeline}. For the baseline model, training with the real-world \datasetname{} dadataset yields limited performance, while introducing the ISP-based simulation (ISPS) significantly improves all metrics, as is presented in Tab.~\ref{tab:simulation ablation}. Combining \datasetname{} with ISPS further brings consistent gains, achieving the best results across all metrics. This indicates that more realistic and diverse degradation modeling is crucial for improving robustness and perceptual fidelity.

\noindent\textbf{Model Design.} Results in Tab.~\ref{tab:model ablation} indicate that adding FS module leads to clear performance improvements, demonstrating effectiveness of frequency-aware modeling. Further incorporating TA loss consistently boosts performance on all metrics, resulting in the best results. These experiments confirm that each component contributes positively, and their combination is essential for achieving optimal performance in the joint flicker-banding and moiré restoration.

\newpage

\section{Conclusion}

In this work, we systematically investigate the problem of image restoration under the coexistence of flicker-banding and moiré artifacts, which frequently occur in real-world screen-captured images but are rarely addressed jointly. So we construct a real-world joint flicker-banding and moiré dataset. Through extensive experiments on our proposed dataset, we find that existing methods are insufficient for this compound scenario, and that simply cascading strong debanding and demoiré models leads to suboptimal results. To overcome these limitations, we propose CLEAR, a unified end-to-end framework that jointly performs debanding and demoiré. CLEAR consistently achieves superior quantitative performance and delivers visually cleaner results with fewer residual artifacts and better structural fidelity. Overall, CLEAR provides a practical and effective solution for real-world flicker–banding \& moiré restoration.

\newpage
\section*{Impact Statement}

This paper presents work whose goal is to advance the field of Machine Learning. There are many potential societal consequences of our work, none which we feel must be specifically highlighted here.  

\bibliography{example_paper}
\bibliographystyle{icml2026}

%%%%%%%%%%%%%%%%%%%%%%%%%%%%%%%%%%%%%%%%%%%%%%%%%%%%%%%%%%%%%%%%%%%%%%%%%%%%%%%
%%%%%%%%%%%%%%%%%%%%%%%%%%%%%%%%%%%%%%%%%%%%%%%%%%%%%%%%%%%%%%%%%%%%%%%%%%%%%%%
% APPENDIX
%%%%%%%%%%%%%%%%%%%%%%%%%%%%%%%%%%%%%%%%%%%%%%%%%%%%%%%%%%%%%%%%%%%%%%%%%%%%%%%
%%%%%%%%%%%%%%%%%%%%%%%%%%%%%%%%%%%%%%%%%%%%%%%%%%%%%%%%%%%%%%%%%%%%%%%%%%%%%%%
% \newpage
% \appendix
% \onecolumn

% You can have as much text here as you want. The main body must be at most $8$
% pages long. For the final version, one more page can be added. If you want, you
% can use an appendix like this one.

% The $\mathtt{\backslash onecolumn}$ command above can be kept in place if you
% prefer a one-column appendix, or can be removed if you prefer a two-column
% appendix.  Apart from this possible change, the style (font size, spacing,
% margins, page numbering, etc.) should be kept the same as the main body.
%%%%%%%%%%%%%%%%%%%%%%%%%%%%%%%%%%%%%%%%%%%%%%%%%%%%%%%%%%%%%%%%%%%%%%%%%%%%%%%
%%%%%%%%%%%%%%%%%%%%%%%%%%%%%%%%%%%%%%%%%%%%%%%%%%%%%%%%%%%%%%%%%%%%%%%%%%%%%%%

\end{document}